\documentclass[letterpaper, 10 pt, conference]{ieeeconf}  

\IEEEoverridecommandlockouts                              

\title{\LARGE \bf
DexTele: A Dual-Arm Dexterous Teleoperation System Based on Motion Retargeting and Adaptive Force Control}

\author{Yuanchuan Lai$^{1}$, Qing Gao$^{1,*}$, Ziyan Liang$^{1}$, Xianfeng Cheng$^{1}$, Junjie Hu$^{2}$, Zhaojie Ju$^{3}$
\thanks{This work was supported in part by the Guangdong Basic and Applied Basic Research Foundation under Grant 2025A1515011954, 2023A1515110074, in part by the Shenzhen Science and Technology Program under Grant ZDCY20250901100201002.}
\thanks{$^{1}$Yuanchuan Lai, Qing Gao, Ziyan Liang and Xianfeng Cheng are with the School of Electronics and Communica-tion Engineering, Sun Yat-sen University, Shenzhen 518107, China.(email:laiych25@mail2.sysu.edu.cn, gaoqing2@mail.sysu.edu.cn)}%
\thanks{$^{2}$Junjie Hu is with the School of Artificial Intelligence, The Chinese University of Hong Kong, Shenzhen, Shenzhen 518172, China.(email: hujunjie@cuhk.edu.cn
)}%
\thanks{{$^{3}$Zhaojie Ju is with the School of Computing, University of Portsmouth, Portsmouth PO1 3HE, UK.(email: Zhaojie.Ju@port.ac.uk)}%
}
\thanks{$^{*}$Corresponding Author:Qing Gao, gaoqing2@mail.sysu.edu.cn.}%
}
\usepackage{soul} 
\usepackage{color, xcolor} 
\usepackage{cite}
\usepackage{graphicx}
\usepackage[colorlinks,bookmarksopen,bookmarksnumbered,citecolor=green, linkcolor=red, urlcolor=green]{hyperref}
\usepackage{booktabs}
\usepackage{multirow}
\usepackage{amsmath}
\usepackage{amssymb}   

\hypersetup{
    colorlinks=true,
    urlcolor=magenta,
}

\begin{document}

\maketitle
\pagestyle{empty}  
\thispagestyle{empty}
\pagestyle{empty}

\begin{abstract}
In dual-arm dexterous teleoperation, cross-platform generalization of motion retargeting and interactivity of grasping are crucial. However, the heterogeneity of robotic architectures and the wide variety of grasping objects pose significant challenges to achieving precise motion retargeting and compliant grasping in dual-arm dexterous teleoperation. To address these challenges, a dual-arm dexterous teleoperation system (DexTele) is proposed based on motion retargeting and adaptive force control. First, a vision-based motion retargeting module is designed to generate preliminary robot motions from human images. In this module, a motion-graph encoder and latent optimization are proposed for precise and convenient cross-platform motion retargeting. Second, an adaptive grasping module is designed to achieve compliant grasping. This module combines a vision-language model (VLM) with model predictive control (MPC), allowing the system to predict the required grasping force for a target object and perform gradient-based online optimization. Finally, extensive experiments demonstrate that the DexTele achieves precise motion retargeting and compliant grasping with generalization across multiple robot platforms. Project can be found at: \href{https://3469627147abc.github.io/DexTele/}{https://github.io/DexTele.}

\end{abstract}

\section{Introduction}
Robotic teleoperation enables human operators to control robots remotely for complex tasks. A critical component is motion retargeting, which maps human movements to robots for natural and precise reproduction. Existing systems often implement motion retargeting through direct mappings, which perform adequately for simple tasks \cite{a4,a5}. However, the limitations of these approaches become evident when extending them to multiple robotic platforms. They are typically designed for a single platform and lack cross-platform generalization, which can easily lead to motion retargeting inaccuracies and compromise the naturalness and reliability of operations on robots with different architectures. In dexterous hand teleoperation, hand motions must not only replicate human gestures but also adapt to object interaction characteristics to ensure precise and safe manipulation \cite{a6,a7}. In this context, adaptive grasping is particularly important, as it enables the robot to adjust its actions based on object properties, thereby ensuring stability and safety during manipulation. Overall, cross-platform generalization, retargeting accuracy, and adaptive grasping jointly determine the practicality and reliability of teleoperation systems.

\begin{figure}
\includegraphics[width=\linewidth]{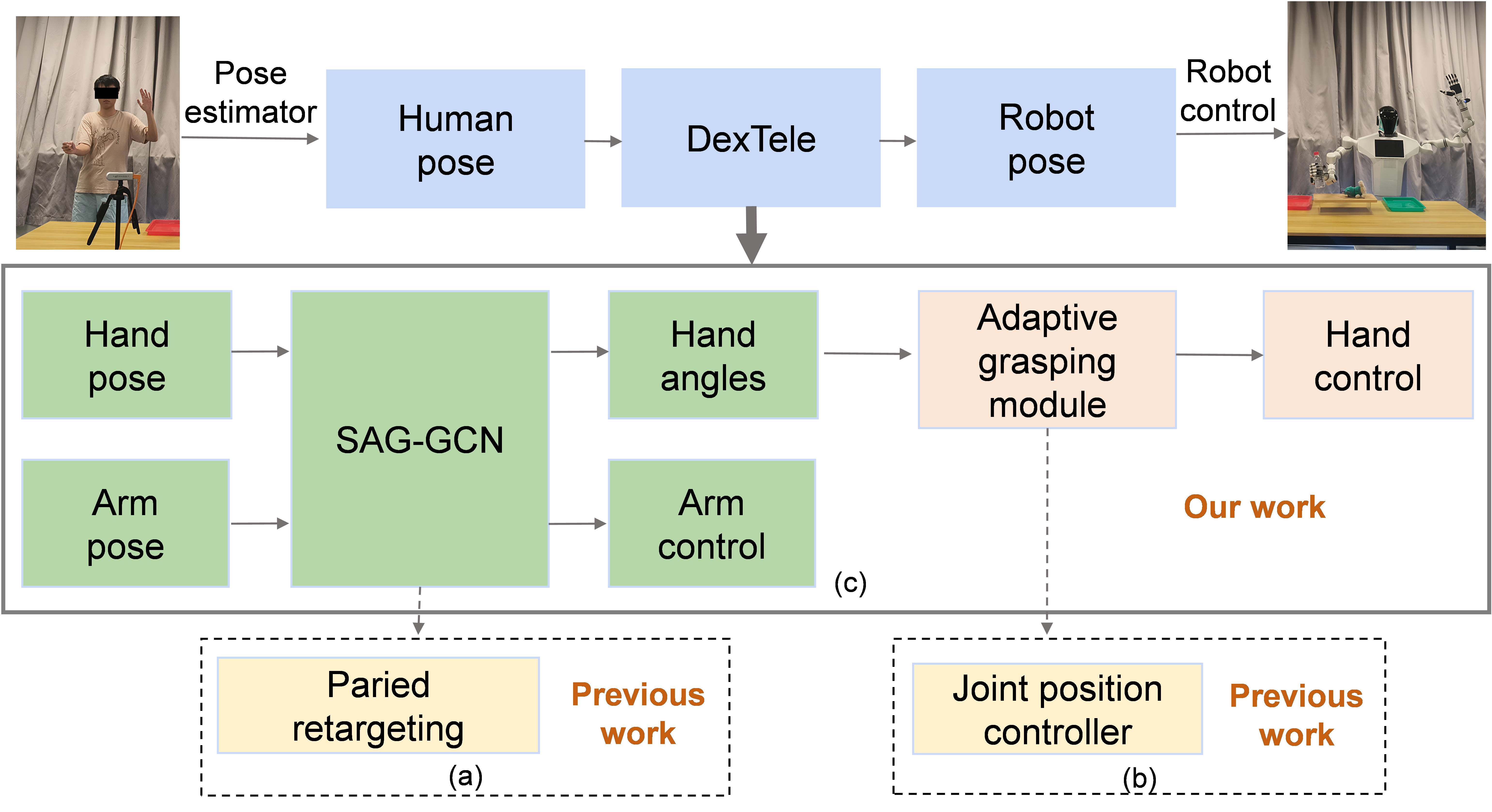}
\centering
\caption{Schematic of the teleoperation system. Part (a) illustrates previous work on motion retargeting, part (b) illustrates previous work on force control, and part (c) presents the proposed pipeline for dexterous teleoperation.} \label{figbegin}
\end{figure}

As illustrated in Fig.\ref{figbegin} (a), existing retargeting methods typically rely on paired human–robot datasets to train supervised mapping models \cite{a8,a9,c1}. Although this approach can improve accuracy in specific scenarios, its limitations are evident: adding a new robot platform requires the collection of additional paired data, thereby constraining the method’s generalization capability. For hand teleoperation, mainstream methods often use position control or fixed force thresholds \cite{a11,a12,c2} (Fig.\ref{figbegin} (b)), which lack the flexibility to handle diverse objects. Consequently, two major challenges arise: retargeting methods struggle to generalize across robot platforms, and hand control methods fail to provide adaptive grasping for diverse objects. Addressing these limitations necessitates a unified teleoperation system that enables efficient cross-platform retargeting while supporting adaptive grasping for dexterous manipulation.
To this end, two key directions emerge as potential solutions. First, cross-platform retargeting can be reformulated as a graph-based cross-topology mapping problem, where structural relationships between humans and robots are represented on a unified graph. The resulting features are embedded into a latent space for optimization, which relies only on human motion data and thus enables motion retargeting across different robot platforms without additional paired datasets. Second, hand grasping is elevated from fixed-threshold control to intelligent strategies that integrate semantic reasoning and dynamic optimization. A VLM understands object characteristics, and MPC generates proactive force control, enabling stability and adaptivity for dexterous hands in complex object interaction scenarios.

Based on aforementioned viewpoints, we propose DexTele, as depicted in Fig.\ref{figbegin} (c). DexTele employs a Spatial Attention Gated Graph Convolutional Network (SAG-GCN) with a dual-stream input–output design, which independently processes arm and hand motions while sharing intermediate representations, thereby improving retargeting accuracy across different robot platforms. Simultaneously, the adaptive grasping module utilizes a VLM to infer object categories and recommend appropriate grasping forces, with subsequent integration into MPC for online optimization. This generates adaptive grasping strategies that balance safety, stability, and foresight. Through extensive simulations and real-world experiments, DexTele has been shown to achieve precise motion retargeting and compliant grasping, with generalization across multiple robot platforms.

Our contributions are summarized as follows:
\begin{itemize}
\item We propose DexTele to address cross-platform motion retargeting and adaptive grasping in teleoperation. It enables precise motion retargeting across robots while supporting stable and flexible object grasping, addressing limitations of existing methods.
\item We propose a vision-based motion retargeting module that uses an SAG-GCN to model human–robot topology and a dual-stream input–output design for arm and hand motions, enabling precise cross-platform motion retargeting.
\item We propose an adaptive grasping module to enable compliant grasping. By integrating a VLM with MPC, it predicts appropriate grasping forces and performs online optimization for stable and adaptive object grasping.

\end{itemize}

\section{Related Work}

\subsection{Human-to-Robot Motion Retargeting}
Human–robot motion retargeting aims to map human movements onto robots with differing kinematics and degrees of freedom for high-fidelity reproduction. Traditional motion capture methods, such as VR headsets, data gloves, or marker-based optical systems \cite{a16, a17}, offer high accuracy but are bulky, costly, and limit user comfort. To overcome these issues, vision-based, non-contact approaches have gained attention for their low cost and ease of deployment \cite{a18}. Following this trend, this study employs a standard RGB camera and the FrankMocap \cite{a37} algorithm, a 3D human pose estimator, to capture human arm–hand poses for robot motion generation.

Current human–robot motion retargeting has mostly focused on individual body parts, such as the hand or arm \cite{a20,a23}, limiting practical applications. Some studies explored joint arm–hand retargeting using kinematics-based methods, offering cross-platform adaptability but limited high-fidelity reproduction. Li et al. \cite{a21} developed a kinematics-driven arm retargeting method for platform adaptation, while Qin et al. \cite{a22} proposed a similar approach with comparable precision limitations. Other approaches rely on paired human–robot datasets. Zeng et al. \cite{a26} developed a teleoperation system with adaptive force control, effective within the trained platform, while Li et al. \cite{a27} proposed a vision-based end-to-end framework for the Shadow hand, accurate but robot-specific. 

To overcome these limitations, SAG-GCN encodes human motion and URDF-based robot models as motion graphs with latent optimization. It further adopts a dual-stream input–output design, processing arms and hands independently while sharing intermediate representations to achieve accurate and scalable retargeting across multiple robot platforms.

\subsection{Force-Feedback-Based Adaptive Dexterous Grasping}
Dexterous grasping and manipulation in complex environments are central to human-level robotic manipulation. However, the diversity of object materials, shapes, and masses makes it difficult to determine suitable grasping forces and adjust them dynamically during execution, which remains a core challenge for adaptive grasping. Traditional methods often employ position control or open-loop force control strategies \cite{a28,a29}, which can accomplish basic pick-and-place tasks but struggle to ensure safe handling of fragile or deformable objects and adapt to external disturbances. Some studies use force sensors for closed-loop control or impedance control to enhance stability \cite{a32},\cite{a33}, but these approaches typically require precise modeling of hand–object interaction dynamics, making deployment complex and limiting generalization in multi-object scenarios.

Recent advances combine machine learning with MPC to enable force prediction and regulation by learning mappings between joint angles and contact forces. For example, Xu et al. \cite{a34} employ GelSight tactile feedback for online adjustment, Shi et al. \cite{a35} integrate Gaussian process-based modeling of state and force relationships into a safety-constrained MPC framework, and Tian et al. \cite{a36} combine deep reinforcement learning with force feedback for multi-finger adaptive grasp control. While these methods provide online force control and prediction, most rely on offline training or lack intelligent perception of target forces and task-specific adaptability.

In this paper, a force-adaptive grasping strategy is introduced to address these difficulties, integrating VLM inference with MPC-based force regulation. By combining VLM task inference with MPC’s real-time optimization, the approach achieves smooth and adaptive grasping across diverse and unseen objects.

\begin{figure*}
\includegraphics[width=\linewidth]{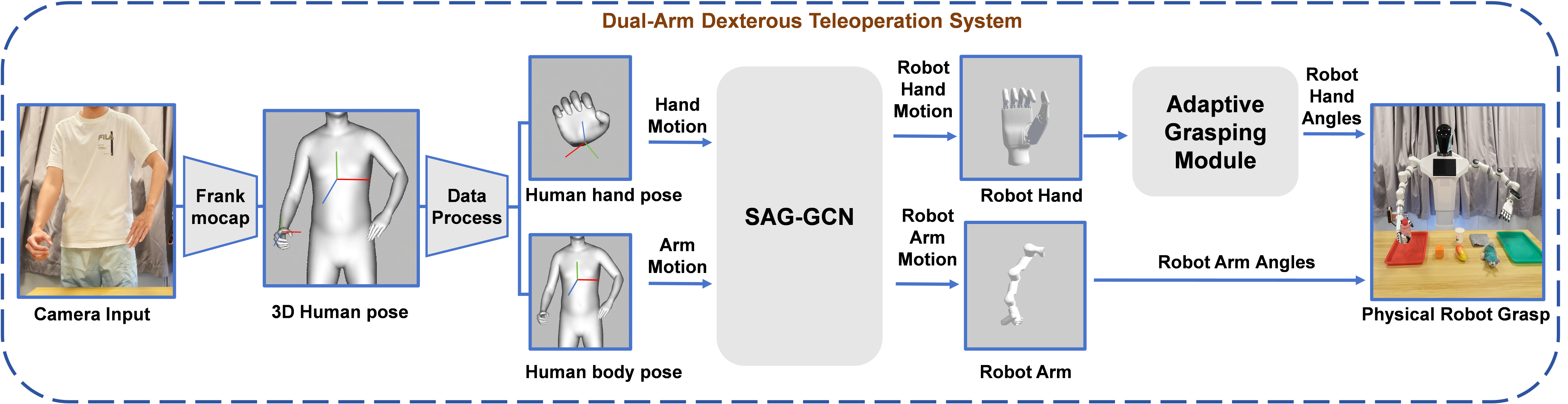}
\centering
\caption{Overview of the dual-arm dexterous teleoperation system. Human motions are first captured via FrankMocap and processed into 3D body and hand poses. The extracted hand and arm motions are retargeted to the corresponding movements of the robot's hand and arm. The robot's arm angles are directly mapped to the robot, while the hand angles are adjusted by the adaptive grasping module before executing the physical grasping.} \label{fig1}
\end{figure*}

\section{Dual-Arm Dexterous Teleoperation System}
\subsection{Overview of DexTele}
This study introduces DexTele, a dual-arm dexterous teleoperation system that integrates vision-based motion retargeting with adaptive grasping. The overall workflow is illustrated in Fig. \ref{fig1}. Motion capture is performed using FrankMocap to obtain three-dimensional human arm and hand motion data. After data processing, the captured motion is separated into independent arm and hand motions due to their distinct characteristics.

The motions are processed by SAG-GCN, with input–output stages handling arms and hands independently. Tailored optimization is then applied to each motion scale and type to enhance retargeting accuracy. The retargeted joint angles of the robot arm are directly executed for control, while those of the dexterous hand are fed into an adaptive grasping module that combines VLM inference with MPC-based force regulation. During grasping, images captured by an external camera are analyzed by the VLM to infer the required grasping force for the target object. This estimate is refined by integrating real-time force feedback from the dexterous hand and applying MPC to dynamically optimize joint commands and applied forces.

\subsection{Motion Retargeting}
\textbf{\textit{Motion Retargeting Problem Statement}}: Motion retargeting is formulated as a latent-space optimization problem. Given a sequence of human skeletal graphs $D={G_k}$, an encoder $f_\phi$ maps $G_k$ into a latent vector $z$, and a decoder $f_\psi$ produces robot joint angles $\theta$, which are further converted to the retargeted trajectory $S$ via forward kinematics $K(\cdot)$. The objective is to minimize the retargeting loss, while the constraint $ \theta_{lower}$ and $\theta_{upper}$ ensure that the predicted joint angles stay within their mechanical limits:
\begin{equation}
\theta=\min_{\phi,\psi} L_{ret}(D,S=K(\theta)),
\end{equation}
\begin{equation}
s.t.{{\theta }_{lower}}\le \theta \le {{\theta }_{upper}}.
\end{equation}

To quantify the discrepancy between the retargeted robot motion and the target human demonstration, we define a composite objective function:
\begin{equation}
\begin{array}{l}
{{L}_{ret}}={{\lambda }_{ee}}{{L}_{ee}}+{{\lambda }_{ori}}{{L}_{ori}}+{{\lambda }_{norm}}{{L}_{norm}}
\\{\rm{                }}\quad\quad\quad\quad+{{\lambda }_{d}}{{L}_{d}}+\lambda_{fin1} L_{fin1} + \lambda_{fin2} L_{fin2}.
\end{array}
\end{equation}

Here, the terms correspond to: end-effector position loss $L_{ee}$, end-effector orientation loss $L_{ori}$, arm normal vector loss $L_{norm}$, dynamics loss $L_{d}$, fingertip orientation loss $L_{fin1}$, and finger angle loss $L_{fin2}$. The associated weights $\lambda_{ee}$, $\lambda_{ori}$, $\lambda_{norm}$, $\lambda_{d}$, $\lambda_{fin1}$, and $\lambda_{fin2}$ are set to 1000, 100, 1000, 1000, 100, and 100, respectively, balancing the contribution of each term to ensure accurate, natural, and physically plausible retargeting.

\begin{figure}
\includegraphics[width=\linewidth]{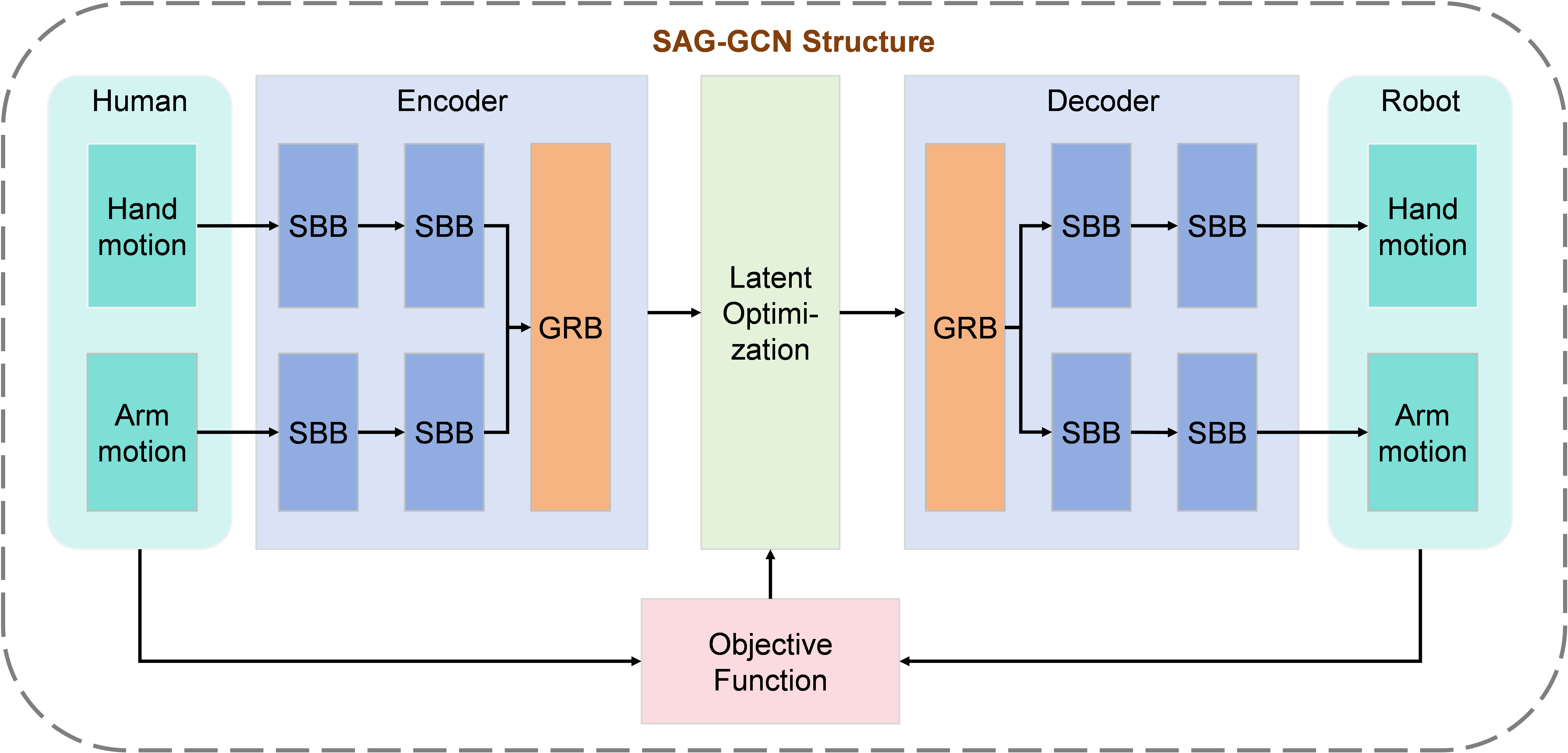}
\centering
\caption{Overview of the SAG-GCN Structure. Human hand and arm motions are captured and processed through the encoder, which utilizes spatial basic blocks and gated residual blocks for feature extraction. The extracted features are then optimized in the latent space using objective functions. Finally, the decoder translates the optimized representations into the corresponding robot joint space, facilitating effective motion retargeting.} \label{fig3}
\end{figure}

\textbf{\textit{Motion Retargeting Architecture}}: The proposed motion retargeting network architecture features a symmetric encoder-decoder design, with each component consisting of three layers, as shown in Fig. \ref{fig3}. The encoder's first two layers utilize a dual-stream structure to address scale differences between arm and hand movements, while the third layer is dedicated to fusion processing. The decoder mirrors this structure for efficient feature integration. The dual-stream input-output design separately processes arm and hand motions, capturing coarse-grained arm movements and fine-grained hand movements to optimize feature learning. The third-layer fusion integrates these features, coordinating arm–hand motions to enhance retargeting accuracy and overall task efficiency.

\begin{figure}
\includegraphics[width=\linewidth]{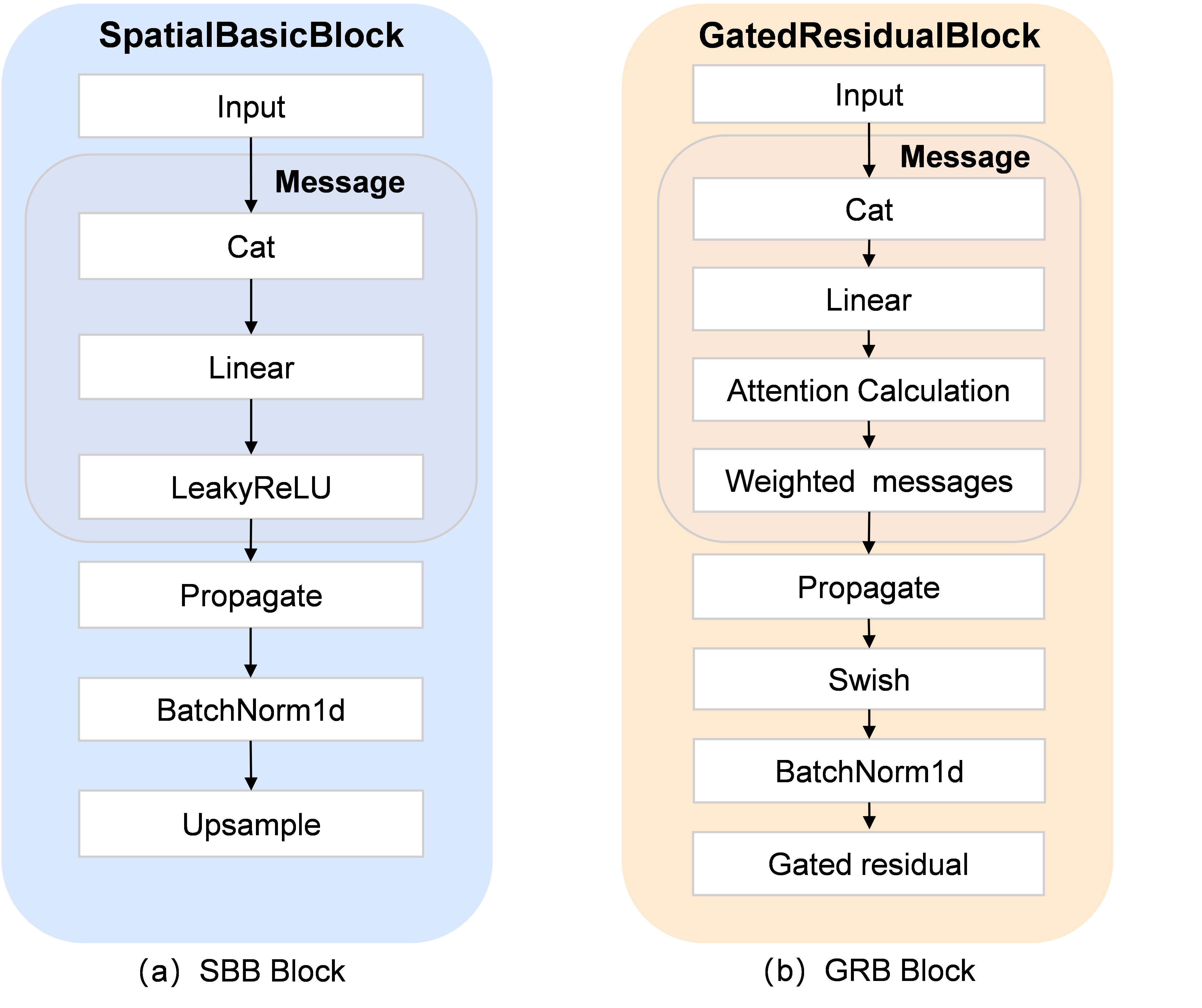}
\centering
\caption{Structures of the Spatial Basic Block and Gated Residual Block.} \label{fig4}
\end{figure}

The core of the retargeting network comprises two modules: the Spatial Basic Block (SBB) and the Gated Residual Block (GRB), with their specific structures illustrated in \ref{fig4}. The SBB acts as a feature extractor, using concatenation and message propagation to process node features and edge attributes, resulting in efficient skeletal topology encoding for real-time motion retargeting. The GRB introduces an attention mechanism and gated residual structure, enhancing feature selectivity and stability while reducing noise accumulation.

\textbf{\textit{Motion Retargeting Network}}: The proposed motion retargeting network uses an end-to-end encoder-decoder architecture with SAG-GCN to map human motion to robot movements. Both human and robot skeletons are represented as weighted graphs to capture joint topology and spatial constraints. At frame $k$, the skeleton is represented as $G_k = (V_k, E_k, W_k)$, where $V_k = {v_{k,1}, \dots, v_{k,N}}$ are joint nodes, $E_k \subseteq V_k \times V_k$ defines connectivity, and $W_k \in {R}^{N\times N}$ is a dynamic attention matrix learned during forward propagation. Each node $v_{k,i}$ carries features $h_{k,i} = [p_{k,i}, q_{k,i}]$ with position $p_{k,i} \in {R}^3$ and quaternion $q_{k,i} \in {R}^4$, while edge features $e_{k,ij} = p_{k,j} - p_{k,i}$ encode local geometry.

During feature encoding, the network addresses scale differences between human and robot skeletons by normalizing positions while keeping rotations unchanged for physical consistency. The normalized position is computed as:
\begin{equation}
\begin{array}{l}
{{\bar{p}}_{k,i}}=\frac{{{p}_{k,i}}-{{c}_{k}}}{{{s}_{k}}},\\{\rm{                }}{{c}_{k}}=\frac{1}{N}\sum\limits_{j=1}^{N}{{{p}_{k,j}}}
,\\{\rm{                }}{{s}_{k}}=\sqrt{\frac{1}{n}\sum\limits_{j=1}^{N}{||{{p}_{k,j}}-{{c}_{k}}||_{2}^{2}}}.
\end{array}
\end{equation}

Here, $c_k$ is the geometric center and $s_k$ the global scale. The normalized positions are concatenated with rotations to form node inputs, which are processed by spatial attention. Attention weights, based on spatial–pose similarity, guide the aggregation of neighborhood information:
\begin{equation}
{{m}_{k,i}}=\sum\limits_{j\in N(i)}{{{\alpha }_{k,ij}}\cdot \phi ([{{{\bar{p}}}_{k,i}},{{q}_{k,i}},{{{\bar{p}}}_{k,j}},{{q}_{k,j}},{{e}_{k,ij}}])},
\end{equation}
where $\phi(\cdot)$ denotes a two-layer fully connected message encoder with Swish activation, and $\alpha_{k,ij}$ is the learned attention coefficient.

The updated node representation is obtained by fusing the aggregated message with the residual features through a gated residual unit:
\begin{equation}
{{g}_{k,i}}=\sigma ({{W}_{g}}{{h}_{k,i}}+{{b}_{g}}) ,
\end{equation}
\begin{equation}
{{{h}'}_{k,i}}={{g}_{k,i}}\odot {{m}_{k,i}}+(1-{{g}_{k,i}})\odot U{{h}_{k,i}},
\end{equation}
where $g_{k,i}$ is the gating vector controlling the fusion ratio, $\sigma(\cdot)$ is the sigmoid function, and $U$ is the dimensional mapping matrix. This gating mechanism mitigates noise accumulation and improves stability in deep feature propagation.

\subsection{Adaptive Grasping}
\begin{figure}
\includegraphics[width=\linewidth]{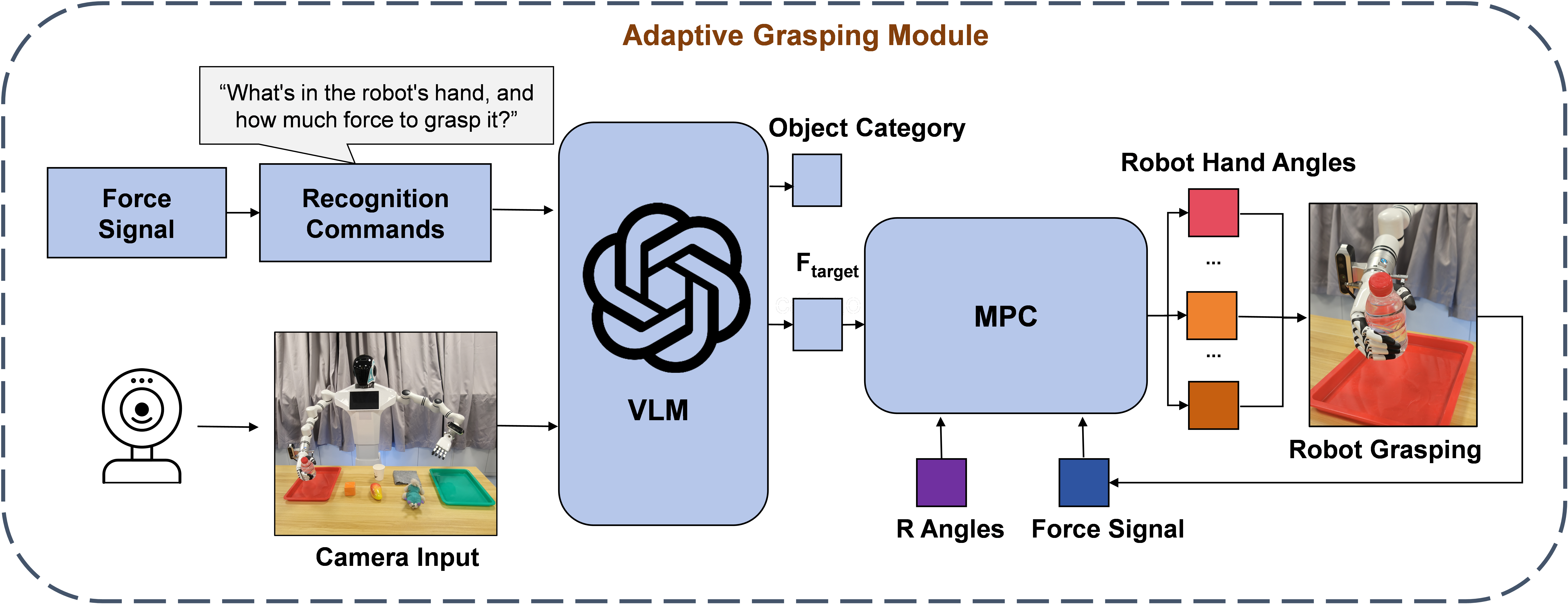}
\centering
\caption{Pipeline of the Adaptive Grasping Module. When the dexterous hand generates force signal, the current image from an external camera is collected and input into the VLM with the recognition commands. The VLM outputs the target grasping force and object category. The MPC module then adjusts the dexterous hand joint angles in real-time to achieve the target force based on the current force signal and retargeted angles (R Angles).} \label{fig5}
\end{figure}

To enable flexible grasping of various objects, an adaptive grasping module has been developed, which combines VLM inference with force regulation based on MPC, as shown in Fig. \ref{fig5}. The core process consists of three consecutive stages: target object recognition, target grasping force estimation, and online optimization of joint commands, thereby establishing a closed-loop mapping from visual perception to adaptive force regulation.

\textbf{\textit{VLM-Driven Target Force Inference}}: At the initial stage of the grasping process, when the dexterous hand generates force signal, an externally mounted camera captures the image of the robot's grasp. The captured image, along with pre-defined recognition commands, is input into the VLM for processing, enabling rapid semantic interpretation and producing two outputs: the object category and the corresponding recommended target grasping force. Compared to traditional methods based on sensors or heuristic force threshold determination, this approach leverages the extensive knowledge base of large models to directly map the object category to the required force. For example, when an object is identified as a "water bottle", the model can infer that the recommended target grasping force is approximately 300g, providing a quantitative reference for subsequent adaptive force optimization.

In this context, the primary advantage of VLM inference lies in its ability to provide a priori estimates of the target force based on knowledge-driven insights. In complex multi-object environments, manual calibration or fixed thresholds often fail to strike a balance between grasping stability and object safety. By utilizing semantic awareness and prior knowledge, the VLM generates reasonable target forces based on visual input, effectively initializing the adaptive force control process.

\textbf{\textit{Joint Angle-Force Prediction Model Construction}}:
After determining the target force, the system estimates the forces acting on the robotic hand from joint angle information to evaluate grasp performance during closed-loop optimization. A data-driven strategy is adopted by training a joint angle–force prediction model on historical motion–force data, represented as:
\begin{equation}
D=\{(\theta _{1}^{k},\theta _{2}^{k},{{F}^{k}})\}_{k=1}^{M},
\end{equation}
where $\theta_{1} \in {R}^{6}$ denotes the commanded joint angles, $\theta_{2} \in {R}^{6}$ denotes the actual joint angle feedback, and $F \in {R}^{6}$ denotes the corresponding six-dimensional force sensor measurements. During model construction, the control and feedback angles are concatenated as input:
\begin{equation}
x=[{{\theta }_{1}},{{\theta }_{2}}]\in {{R}^{12}}.
\end{equation}

The model output is defined as the predicted force $\hat{F} \in {R}^{6}$. A random forest regressor is employed to capture the nonlinear mapping between joint angles and forces while maintaining a balance between training efficiency and inference speed.

Once trained, the model $M$ produces rapid force predictions without reliance on real-time force sensor readings, serving as a differentiable surrogate for the subsequent gradient-based optimization. This surrogate functions as a mechanical approximation model for the MPC optimizer, enabling direct force response inference in the command space and facilitating online force regulation without the need for explicit dynamic modeling.

\textbf{\textit{Adaptive Force Optimization Based on MPC Principles}}: During grasp execution, joint commands are continuously refined based on real-time feedback to ensure smooth convergence of the output force toward the target value. An MPC-inspired online optimization module integrates the force prediction model into a gradient-based iterative loop, forming a lightweight closed-loop control mechanism.

At each control cycle, the previous joint command $\theta_{1}^{{prior}}$ is combined with the current actual joint angles $\theta_{2}$ to obtain the predicted force $\hat{F}$ from the model. The optimization problem is formulated as:
\begin{equation}
L({{\theta }_{1}})=||M({{\theta }_{1}},{{\theta }_{2}})-{{F}_{target}}|{{|}^{2}}+\lambda ||{{\theta }_{1}}-\theta _{1}^{prior}|{{|}^{2}},
\end{equation}
where the first term enforces proximity between predicted and target forces, and the second term regularizes large variations in joint commands. The weight $\lambda$ balances responsiveness and smoothness.

The joint angles ${{\theta }_{1}}$ are treated as differentiable variables and updated through gradient descent using the Adam optimizer:
\begin{equation}
\theta _{1}^{*}=\underset{{{\theta }_{1}}}{\mathop{\arg \min L}}\,({{\theta }_{1}}).
\end{equation}

The resulting $\theta_{1}^{*}$ is applied in the next control step, enabling rolling optimization with MPC-like properties. This approach preserves motion continuity while adaptively compensating for force deviations, ensuring stable and controllable grasping performance.
\section{Experiments}

\subsection{Experimental Setup}
To assess the deployability of the proposed vision-based motion retargeting system across multiple robotic platforms, human motion retargeting experiments were conducted on three robots: RMC-DA, YuMi, and Unitree H1. RMC-DA was evaluated in both physical and virtual environments, while YuMi and Unitree H1 were tested virtually. The RMC-DA platform is equipped with dual arms, each having six degrees of freedom, and force-sensing Inspire Robotics dexterous hands, with each hand possessing 6-DOF. In comparison, the YuMi robot features 7-DOF arms, each mounted with the same dexterous hand as RMC-DA. The Unitree H1 robot employs 5-DOF arms paired with dexterous hands that have 12-DOF.

For training and evaluation, two datasets were utilized. The first is the high-quality open-source human pose dataset Sign \cite{a38}. The second dataset is our custom dataset, constructed from the CSL-Daily sign language image dataset \cite{a39}, utilizing FrankMocap to extract 3D human poses from single-frame images. Both datasets provide positional data and quaternion rotations for three major joints per arm, as well as positional data for sixteen joints per hand. The CSL-Daily-derived dataset captures complex upper-body movements across 151 action classes demonstrated by three individuals, covering a wide range of daily human limb motions.

The graph neural network was implemented with PyTorch Geometric \cite{a40} and trained using the Adam optimizer with a fixed learning rate of 1e-4 on an NVIDIA RTX 4090 GPU and Intel Core i7-11700KF CPU. 

\subsection{Comparative Experiments on Motion Retargeting}

\subsubsection{Arm Motion Retargeting}
Comparative experiments on arm motion retargeting were conducted using the Sign and CSL-Daily datasets, evaluating our method against three baselines: NLO \cite{a38}, VMR \cite{a44}, and ATP \cite{a45}. Four evaluation metrics were adopted for arm motion: Mean Per Joint Position Error (MPJPE) \cite{a41}, Quaternion Distance (Quat) \cite{a42}, Velocity Error (VE), and Acceleration Error (AE) \cite{a43}. MPJPE measures the average distance between predicted and ground-truth joint positions, while Quat assesses the rotational difference in joint orientations. VE and AE evaluate the temporal smoothness of motion through the first- and second-order derivatives of joint positions. As shown in Tables \ref{tab:sign_results} and \ref{tab:csl_results}, our method outperformed all metrics on both datasets, demonstrating higher positional accuracy, improved rotational consistency, and smoother motion dynamics. 

\begin{table*}[t]
\centering
\caption{Evaluation of Arm and Finger Motion Retargeting Performance on the Sign Dataset.}
\label{tab:sign_results}
\begin{tabular}{lccccccccc}
\toprule
{Method} 
& \multicolumn{4}{c}{Arm Metrics} 
& \multicolumn{5}{c}{Finger Metrics} \\
\cmidrule(lr){2-5} \cmidrule(lr){6-10}
& {MPJPE (m)} & {Quat (rad)} & {VE (m/s)} & {AE (m/s$^2$)} 
& {Thumb (rad)} & {Index (rad)} & {Middle (rad)} & {Ring (rad)} & {Pinky (rad)} \\
\midrule
NLO\cite{a38}   & 0.0948 & 0.1670 & 0.0590  & 2.1420 & 0.2200 & 0.2542 & 0.0863 & 0.1609 & 0.1289 \\
VMR\cite{a44}   & 0.0853 & 0.1578 & 0.0358  & 1.1056 & 0.2133 & 0.2631 & 0.0811 & 0.1571 & 0.1263 \\
ATP\cite{a45}   & 0.1021 & 0.1834 & 0.0425  & 2.4312 & 0.2046 & 0.2749 & 0.1034 & 0.1497 & 0.1351 \\
Ours  & \textbf{0.0785} & \textbf{0.1503} & \textbf{0.0304} & \textbf{0.8212} 
      & \textbf{0.1967} & \textbf{0.2471} & \textbf{0.0732} & \textbf{0.1476} & \textbf{0.1223} \\
\bottomrule
\end{tabular}
\end{table*}

\begin{table*}[t]
\centering
\caption{Evaluation of Arm and Finger Motion Retargeting Performance on the CSL-Daily Dataset}
\label{tab:csl_results}
\begin{tabular}{lccccccccc}
\toprule
{Method} 
& \multicolumn{4}{c}{Arm Metrics} 
& \multicolumn{5}{c}{Finger Metrics} \\
\cmidrule(lr){2-5} \cmidrule(lr){6-10}
& {MPJPE (m)} & {Quat Error (rad)} & {VE (m/s)} & {AE (m/s$^2$)} & {Thumb (rad)} & {Index (rad)} & {Middle (rad)} & {Ring (rad)} & {Pinky (rad)} \\
\midrule
NLO\cite{a38}   & 0.1038 & 0.1771 & 0.0622 & 1.9560 & 0.2200 & 0.2542 & 0.0863 & 0.1609 & 0.1289 \\
VMR\cite{a44}   & 0.0963 & 0.1622 & 0.0471 & 1.3296 & 0.2133 & 0.2631 & 0.0811 & 0.1571 & 0.1263 \\
ATP\cite{a45}    & 0.9381 & 0.1778 & 0.0527 & 1.6319 & 0.2169 & 0.2566 & 0.0901 & 0.1568 & 0.1369 \\
Ours  & \textbf{0.0882} & \textbf{0.1550} & \textbf{0.0388} & \textbf{0.9421} & \textbf{0.1833} & \textbf{0.2182} & \textbf{0.0766} & \textbf{0.1385} & \textbf{0.1310} \\
\bottomrule
\end{tabular}
\end{table*}

\subsubsection{Hand Motion Retargeting}
The datasets and baselines for hand motion retargeting are consistent with those for arm motion retargeting. Performance was measured using finger joint angle errors (Fin Angle), computed with the three-point method \cite{a4}. As shown in Tables \ref{tab:sign_results} and \ref{tab:csl_results}, the proposed method consistently achieved the lowest errors across all fingers and datasets. These results confirm that the method provides more precise and consistent retargeting of diverse hand motions.

\begin{table}[t]
\centering
\caption{Ablation study of motion retargeting}
\label{tab:dual_stream_ablation}
\begin{tabular}{lcccc}
\toprule
{Method} & {MPJPE(m)} & {Quat(rad)} & {Fin Angle (rad)}  \\ 
\midrule
Ours (Full)     & \textbf{0.0882} & \textbf{0.1550} & \textbf{0.1495}  \\
Single Graph    & 0.1333 & 0.1821 & 0.1947  \\
w/o GRB         & 0.1513 & 0.1939 & 0.2020  \\
w/o SBB         & 0.1566 & 0.2035 & 0.2239  \\
w/o Hand        & 0.1139 & 0.1630 & ---     \\
w/o Arm         & ---    & ---    & 0.1729  \\
\bottomrule
\end{tabular}
\end{table}

\begin{figure}
\includegraphics[width=\linewidth]{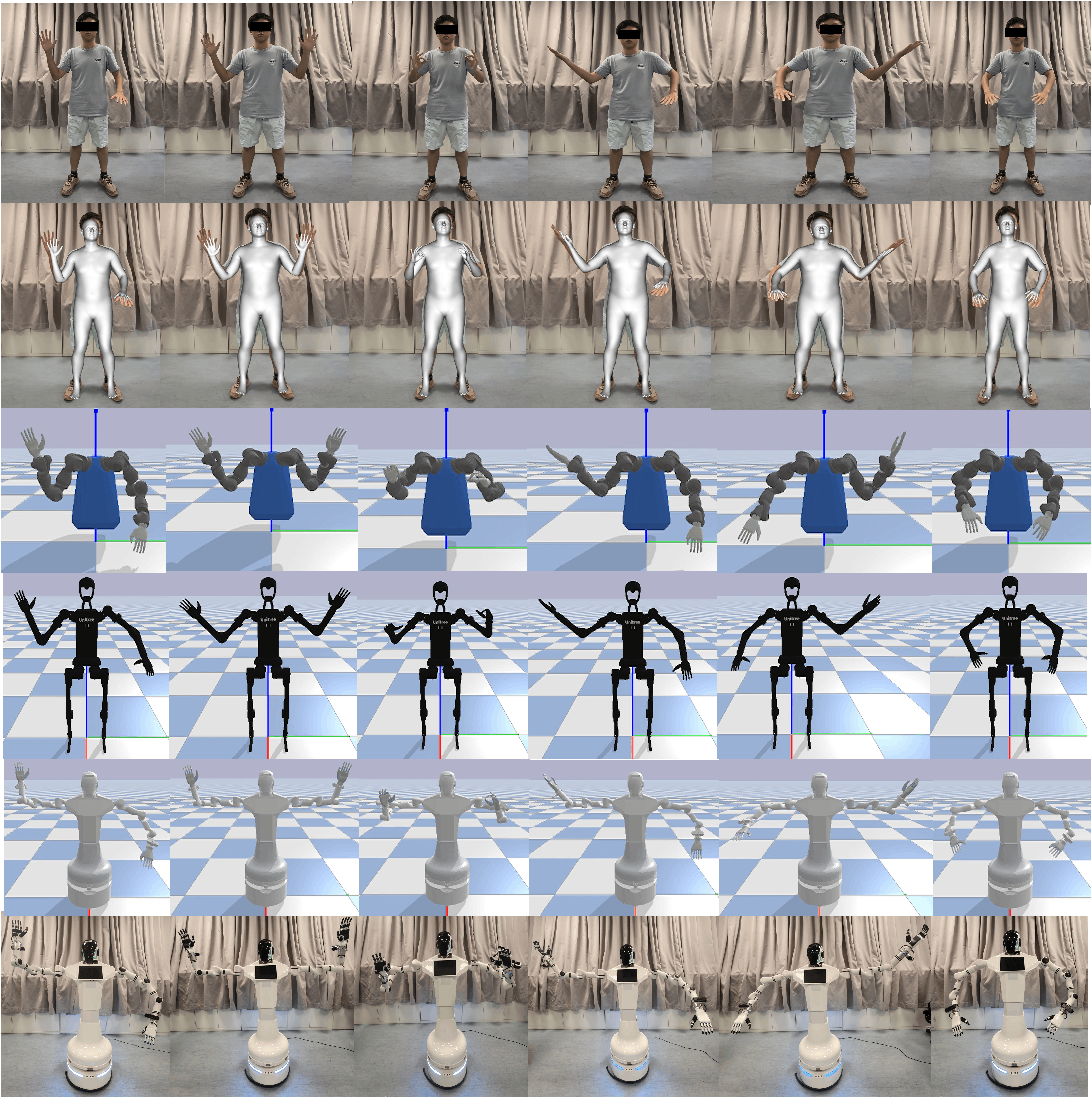}
\centering
\caption{Demonstration of motion retargeting across multiple robot platforms. From top to bottom, the six rows represent: the human demonstrator, the mesh rendering of the human demonstrator, the YuMi robot in simulation, the Unitree H1 robot in simulation, the RMC-DA robot in simulation, and the RMC-DA robot in a real-world environment. }\label{fig6}
\end{figure}

\begin{table}[t]
\centering
\caption{Evaluation of Cross-Platform Robot Performance}
\label{tab:multi_robot}
\begin{tabular}{lcccccc}
\toprule
{Metric} & {YuMi} & {Unitree H1} & {RMC-DA} \\ 
\midrule
{MPJPE (m)}          & 0.1024 & 0.0877 & 0.0882 \\ 
{Quat(rad)} & 0.1735 & 0.1422 & 0.1550 \\ 
{Fin Angle(rad)} & 0.1486 & 0.1845 & 0.1495 \\ 
{Velocity Error (m/s)} & 0.0431 & 0.0622 & 0.0388 \\ 
{Acceleration Error (m/s²)} & 0.8376 & 1.1329 & 0.9421 \\ 
\bottomrule
\end{tabular}
\end{table}

\subsection{Ablation Experiments on Motion Retargeting}
An ablation study on the CSL-Daily dataset evaluated the dual-stream architecture and its key components. Variants included a single-stream structure (Single Graph), removal of the GRB module, replacement of the SBB module with a linear structure, and retaining only the hand or arm stream. As shown in Table~\ref{tab:dual_stream_ablation}, the Single Graph variant introduces interference between local and global features, increasing positional and angular errors. Removing GRB weakens feature fusion and attention, reducing end-effector precision. Replacing SBB with a linear structure limits modeling of nonlinear dependencies, raising MPJPE and Quat errors. Retaining only one stream preserves accuracy for that part but loses complementary information, impairing overall coordination. These results underscore the importance of separately modeling hand and arm motions and integrating them via GRB for coherent and natural retargeting.

\subsection{Multi-Robot Motion Retargeting Experiments}
The cross-platform generalization capability of the proposed vision-guided motion retargeting method was visualized on three robotic platforms: YuMi, Unitree H1, and RMC-DA. As shown in Fig.~\ref{fig6}, a consistent set of human motion samples was applied in both simulated and real environments, demonstrating stable reproduction quality across platforms.

In addition to visualization, we conducted quantitative evaluations on the CSL-Daily dataset using MPJPE, Quat, Fin angle, velocity error, and acceleration error, as summarized in Table~\ref{tab:multi_robot}. All platforms showed low positional and rotational errors, along with smooth motion dynamics, demonstrating the robustness and applicability of the proposed method across multiple robot platforms.

In addition, the performance of real-time teleoperation was evaluated. During teleoperation, the system achieved approximately 10 frames per second, with FrankMocap processing each human pose frame in 0.08 seconds and the retargeting module completing each frame within 0.02 seconds.

\begin{table}[htbp]
\centering
\caption{Adaptive Grasping Performance}
\begin{tabular}{lcccc}
\toprule
{Object} & {VLM Target} & {Force} & {Force} & {Securely } \\ 
                  & {Force(g)}              & {Dev(\%)}            & {Osc
 (\%)}              & {Gripped} \\ 
\midrule
Water Bottle & 300 & 5.3 & 4.2 & \checkmark \\
Beer Cans  & 270  & 7.1 & 6.5 & \checkmark \\
Paper Box          & 60  & 4.9 & 6.1 & \checkmark \\
Plastic Mango      & 150 & 7.6 & 3.2 & \checkmark \\
Sponge Block       & 50  & 6.0 & 2.5 & \checkmark \\
Plush Toy          & 200 & 4.8 & 7.2 & \checkmark \\
Rag        & 50  & 8.1 & 8.4 & \checkmark \\
Paper Cup          & 30  & 5.9 & 7.5 & \checkmark \\

\bottomrule
\end{tabular}
\label{tab:force_feedback}
\end{table}

\begin{table}[htbp]
\centering
\caption{Grasp Success Rate Evaluation}
\begin{tabular}{lcc}
\toprule
Object & Success w/ AGM & Success w/o AGM \\ 
\midrule
Water Bottle    & \textbf{9/10} & 7/10 \\
Beer Cans       & \textbf{9/10} & 6/10 \\
Plastic Mango   & \textbf{10/10} & 4/10 \\
Paper Box       & \textbf{9/10} & 5/10 \\
Sponge Block    & \textbf{8/10} & 4/10 \\
Plush Toy       & \textbf{10/10} & 5/10 \\
Rag             & \textbf{9/10} & 6/10 \\
Paper Cup       & \textbf{9/10} & 4/10 \\
\bottomrule
\end{tabular}

\label{tab:force_feedback_comp}
\end{table}

\subsection{Adaptive Grasping Performance Evaluation}
\subsubsection{Adaptive Grasping Performance}
\begin{figure}
\includegraphics[width=\linewidth]{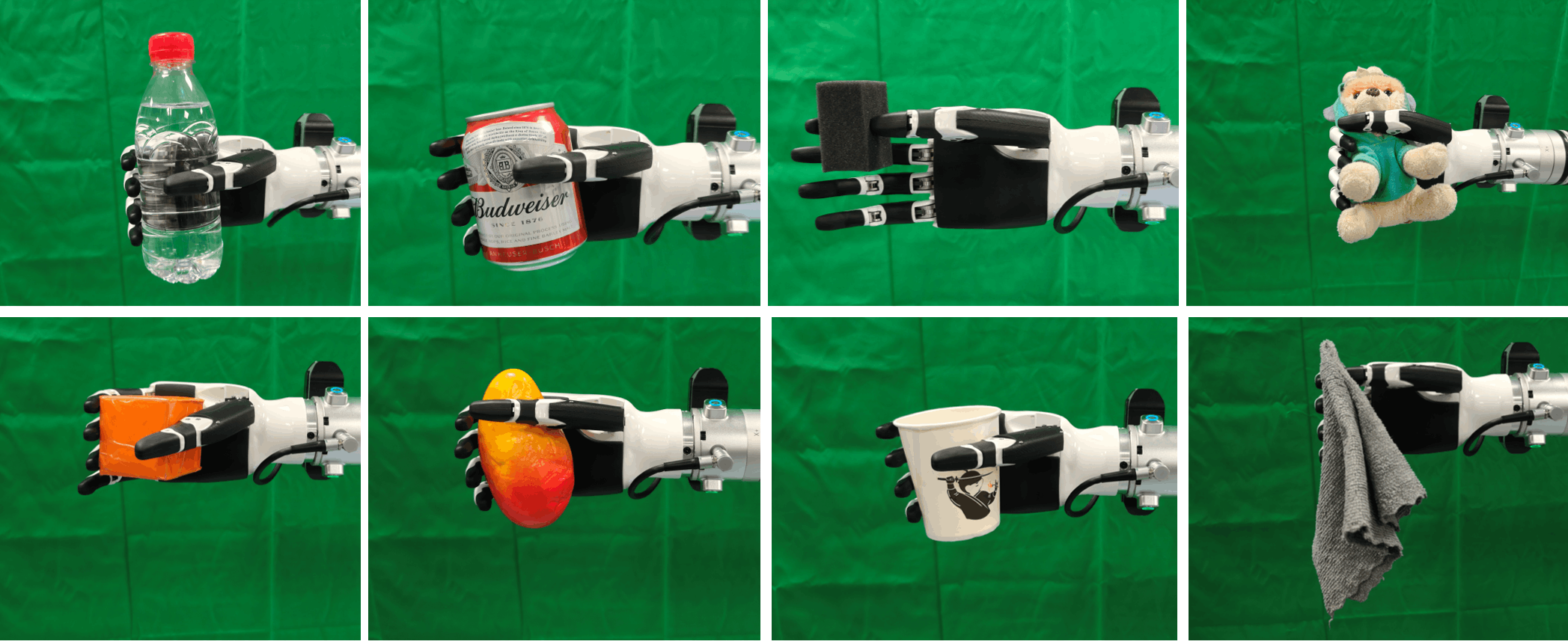}
\centering
\caption{Illustration of the grasping results on eight different object categories, demonstrating the adaptive force control across various textures and shapes.} \label{fig8}
\end{figure}

For VLM selection, the latest Doubao-seed-1-6-vision-250815 model was adopted for visual inference. Under the experimental setup, the average inference latency was approximately 0.2–0.3 s, sufficient to meet the real-time response requirements of robotic grasping. 
The proposed adaptive grasping module has been validated on eight daily objects with varying shapes and materials, with the visual results illustrated in Fig.~\ref{fig8}.

In addition to visualization, the grasping performance on eight object categories was quantitatively evaluated, as summarized in Table~\ref{tab:force_feedback}. The VLM accurately inferred the target grasping force for each object. During grasp execution, both force deviation (Dev) and oscillation (Osc) amplitude remained within 10\% for rigid and deformable objects. All objects were grasped securely without slippage or damage. 

\begin{figure}
\includegraphics[width=\linewidth]{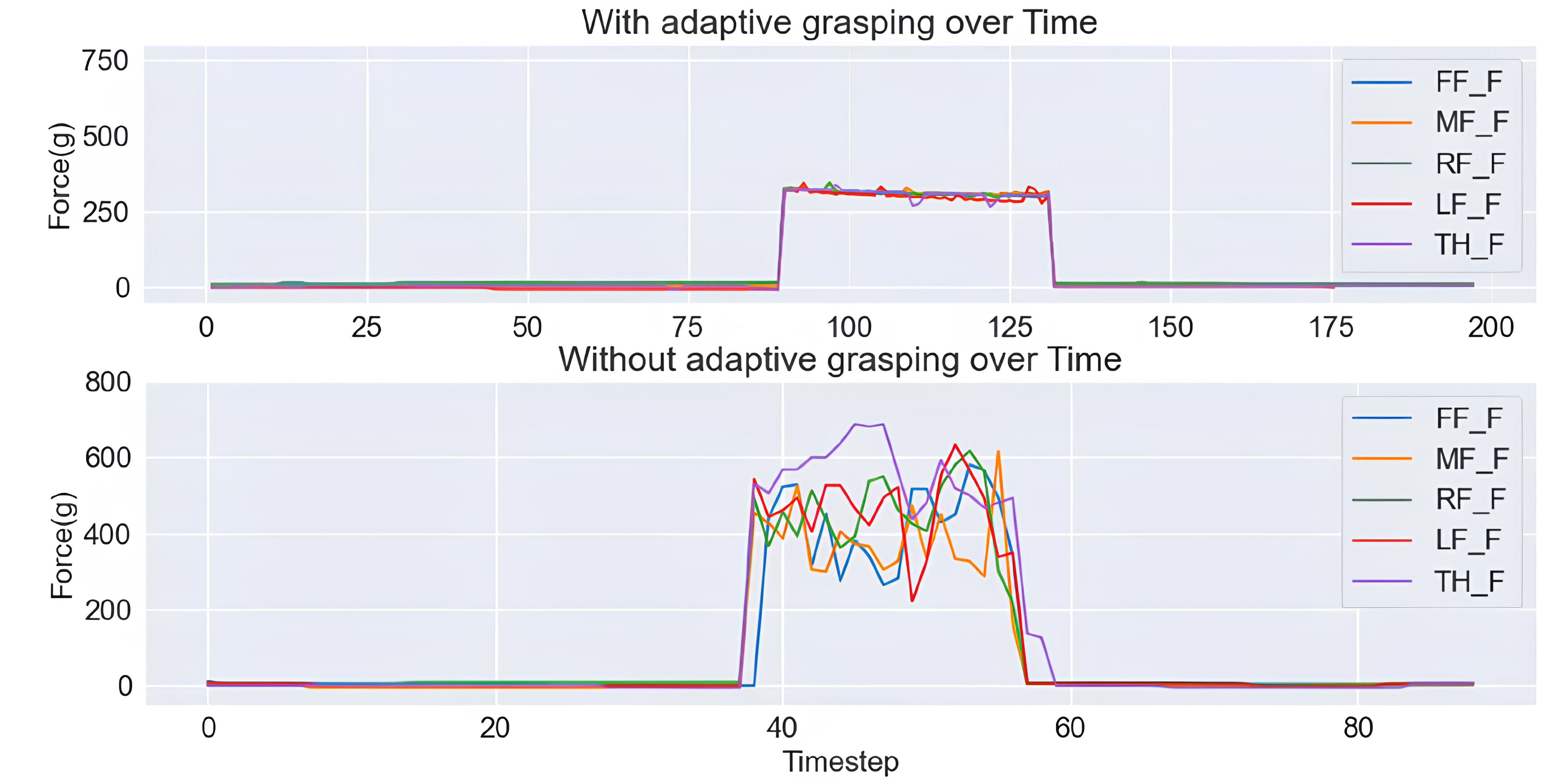}
\centering
\caption{Grasping force curves for the water bottle. The five colored lines correspond to the forces applied by the five individual fingers.} \label{fig7}
\end{figure}

To better illustrate the role of the adaptive grasping module, we visualized the force variation during the process of grasping a water bottle with and without the module, as shown in Fig.~\ref{fig7}. Without the adaptive module, the finger forces remain uncontrolled. In contrast, with the adaptive module, the grasping angles are automatically adjusted to reach the target force and converge smoothly. These results demonstrate that integrating VLM inference with MPC-based force regulation enables precise, low-oscillation, and safe compliant grasping across different types of objects.

\subsubsection{Grasping Success Rate Evaluation}
The capability of the adaptive grasping module (AGM) was evaluated across eight categories of everyday objects, where a successful grasp was defined as a stable hold without slippage or significant deformation. As shown in Table~\ref{tab:force_feedback_comp}, incorporating the AGM substantially improved reliability, with the average success rate increasing from 5.13 to 9.13 out of 10 trials. For deformable objects such as paper cups and plush toys, adaptive force regulation effectively prevented excessive compression and structural damage. For easily slippable items such as plastic mangoes and sponge blocks, rapid adjustment reduced both slippage and unintended release. These findings demonstrate that integrating the proposed AGM into teleoperation can significantly enhance grasp stability and safety.

\section{Conclusion}
This study presents DexTele, which integrates motion retargeting with adaptive force control. Experimental results demonstrate that the proposed vision-based motion retargeting module, implemented with the designed SAG-GCN, enables accurate and efficient cross-platform motion retargeting. In addition, the adaptive grasping module combines a VLM with MPC, allowing the system to infer the required grasping force for target objects and perform gradient-based online optimization to achieve compliant grasping. The system has been successfully validated on multiple robot platforms, including RMC-DA, YuMi, and Unitree H1, showing strong generalization capability and real-time performance.

Nevertheless, certain limitations remain. First, the real-time performance is constrained by the pose estimation algorithm. Second, the retargeting algorithm is currently limited to upper-body robot motion. Future work will focus on adopting more lightweight and accurate pose estimation methods and extending the retargeting algorithm to whole-body robot motion.

\clearpage

\end{document}